\documentclass[runningheads]{llncs}
\usepackage{graphicx}

\usepackage{tikz}
\usepackage{xcolor}
\usepackage{comment}
\usepackage{amsmath,amssymb} 
\usepackage[accsupp]{axessibility}  

\begin{document}
\pagestyle{headings}
\mainmatter
\def\ECCVSubNumber{23}  

\title{LWA-HAND: Lightweight Attention Hand for Interacting Hand Reconstruction} 


\titlerunning{LWA-HAND}
%
\author{Xinhan Di\inst{1} \and
Pengqian Yu\inst{2}}
\authorrunning{X. Di and P. Yu}
%
\institute{\email{xinhan.di@blooxr.com}\\
\and
\email{yupengqian1989@gmail.com}}
\maketitle

\begin{abstract}
Recent years have witnessed great success for hand reconstruction in real-time applications such as visual reality and augmented reality while interacting with two-hand reconstruction through efficient transformers is left unexplored. In this paper, we propose a method called lightweight attention hand (LWA-HAND) to reconstruct hands in low flops from a single RGB image. To solve the occlusion and interaction problem in efficient attention architectures, we propose three mobile attention modules in this paper. The first module is a lightweight feature attention module that extracts both local occlusion representation and global image patch representation in a coarse-to-fine manner. The second module is a cross image and graph bridge module which fuses image context and hand vertex. The third module is a lightweight cross-attention mechanism that uses element-wise operation for the cross-attention of two hands in linear complexity. The resulting model achieves comparable performance on the InterHand2.6M benchmark in comparison with the state-of-the-art models. Simultaneously, it reduces the flops to $0.47GFlops$ while the state-of-the-art models have heavy computations between $10GFlops$ and $20GFlops$.
\end{abstract}

\section{Introduction}
\begin{figure*}
\centering
\includegraphics[height=8cm]{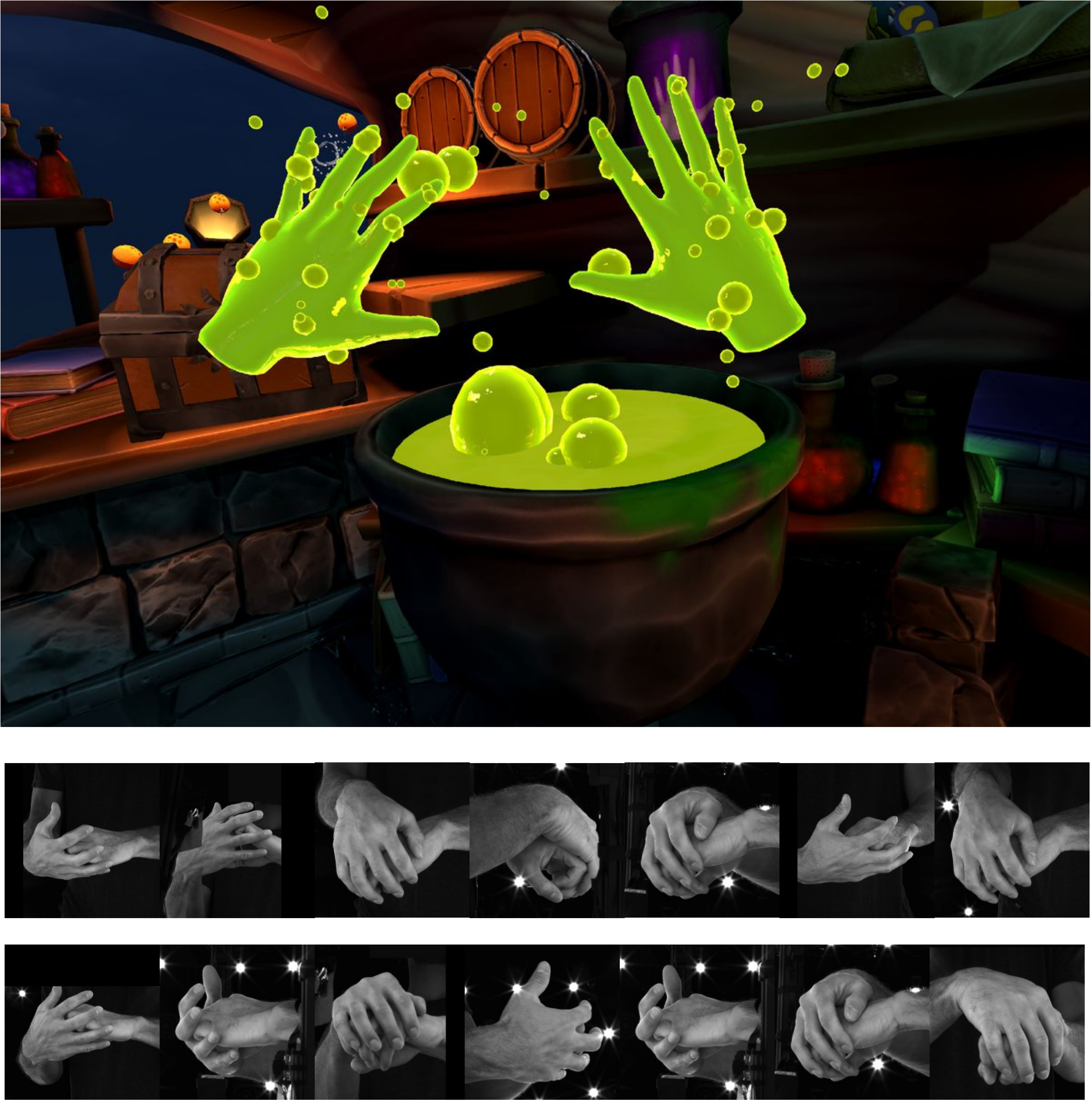}
\caption{Hand tracking application in visual reality. Here interacting hand reconstruction remains  a challenge.}
\label{fig1}
\end{figure*}

Single hand reconstruction has received great success based on deep neural networks \cite{boukhayma20193d,ge20193d,kulon2020weakly,zhang2019end,zhou2020monocular} in industrial applications such as virtual reality (VR), augmented reality (AR), digital shopping, robotics, and digital working. However, interacting two-hand reconstruction is more difficult for real-world applications such as hand tracking in Figure \ref{fig1}. First, it is difficult for networks to obtain the alignment between hand landmarks and image features as feature extractors are confused by mutual occlusions and appearance similarity. In addition, interaction between two-hands is hard to be represented during network training. Finally, it is not trivial to design efficient network architectures which can formulate the occlusion and interaction of two hands and meet the requirement of low latency on mobile hardware at the same time.

Good results have been produced by a variety of monocular double hands tracking techniques based on depth estimation \cite{kyriazis2014scalable,mueller2019real,oikonomidis2012tracking,taylor2017articulated,tzionas2016capturing}. Although these depth-based methods are studied for years, the high computational complexity and the heavy energy consumption limit the application of these methods. Recently, a monocular RGB based reconstruction framework is built through the estimation of a tracking dense matching map \cite{wang2020rgb2hands}. However, its tracking procedure is not able to solve fast motion and the prior knowledge of interacting hands and the context are not fully used through deep networks. Then, the large scale two-hand dataset InterHand2.6M is open \cite{moon2020interhand2}, learning based image reconstruction methods are well explored through the building of different network architectures. Such examples include the 2.5D heatmaps for hand joint positions estimation \cite{fan2021learning,kim2021end,moon2020interhand2} and the attention map for the extraction of sparse and dense image features \cite{zhang2021interacting,li2022interacting}. However, these methods consume a lot of computation power and restrict their applications on mobile devices such as VR/AR glass and robots. In contrast, mobile vision transformers \cite{touvron2021training,mehta2021mobilevit,mehta2022separable,chen2022mobile} has achieved great success in vision tasks and deployment on mobile devices. The lightweight attention based methods are likely to represent the occlusions and interaction of two-hands on mobile devices.

Motivated by the above mentioned challenges, we propose lightweight attention hand (LWA-HAND), a mobile attention and graph based single image reconstruction method. Firstly, two-stream GCN is utilized to regress mesh vertices of each hand in a coarse-to-fine-manner, similar to traditional GCN \cite{ge20193d} and Intaghand \cite{li2022interacting}. However, for the two-hand reconstruction in a low energy consumption, naively using a two-stream network with normal attention modules fails to represent occlusion and interaction of two hands in low latency. Moreover, application of normal feature extraction \cite{he2016deep} and extra attention modules \cite{li2022interacting} for contact image and graph features leads to high flops. To address theses issues, we equip a lightweight vision transformer for feature extraction of interacting hands. This mobile transformer utilizes the data stream of MobileNet \cite{sandler2018mobilenetv2} and transformers with a two-way bridge. This bridge enables bidirectional fusion of local features of each hand and global features of contexts of the occlusions and interaction between hands. Furthermore, a pyramid cross domain module is applied to fuse image representation and hand representation in  a coarse-to-fine and lightweight manner. Global priors of image domain with very few tokens is calculated in a pyramid way and a direct fusion of the prior and  representation of hand is then conducted. Unlike heavy-weight cross hand attention module \cite{li2022interacting}, a separate attention mechanism with linearly complexity \cite{mehta2022separable} reduces the calculation of cross attention to encodes interaction context into hand vertex features. Therefore, the proposed lightweight modules makes the whole network architecture a good choice for resource-constrained devices. Overall, our contributions are summarized as:
\begin{figure*}
\centering
\includegraphics[height=5cm]{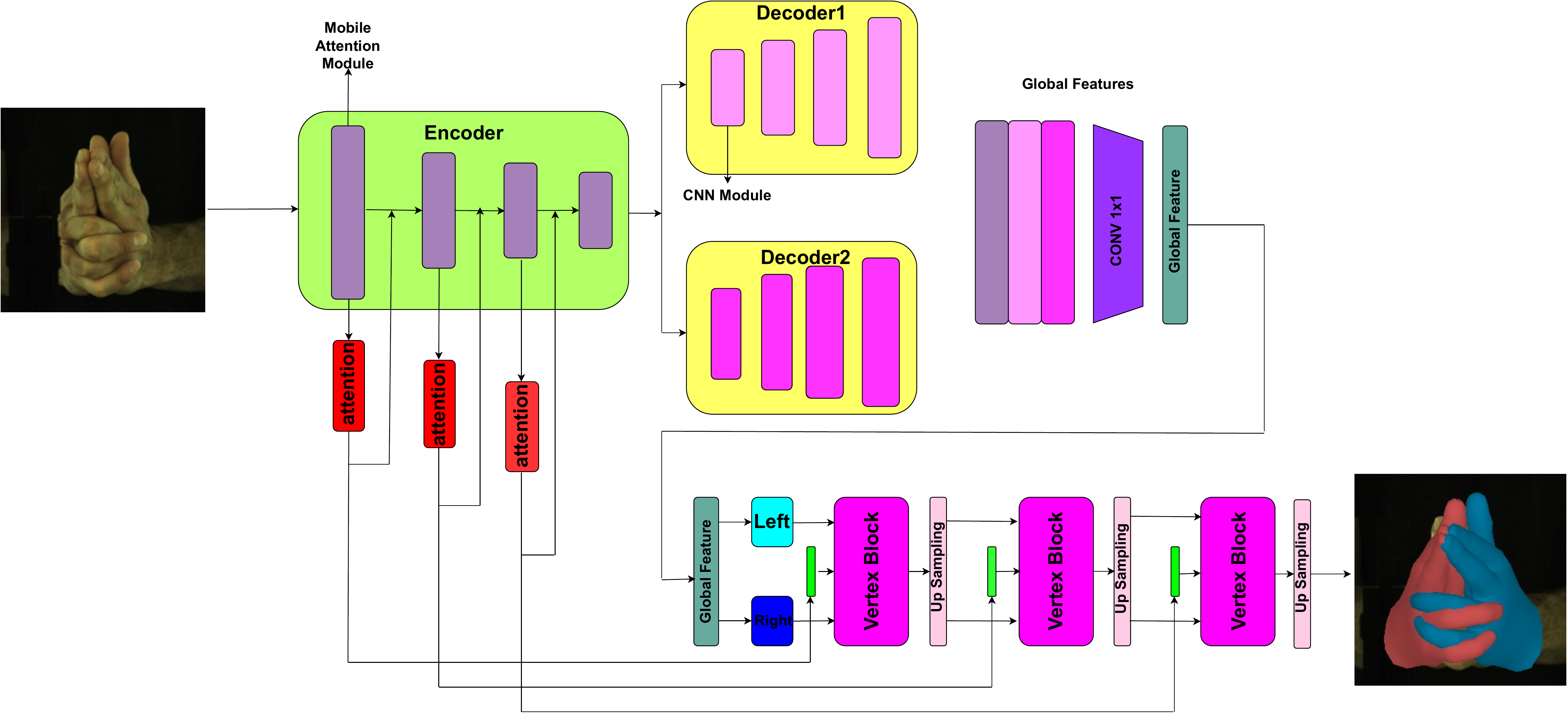}
\caption{Proposed mobile network architecture. The green block is the mobile attention encoder. The two yellow block are decoders of 2.5D heatmap and 2D segmentation. The stacked bridges represent the attention part (in red), the fusion part (in green) and the vertex attention part (in pink), respectively. The mobile attention encoder, stacked bridges, fusion part and vertex attention part are the proposed modules for the efficient hand reconstruction tasks.}
\label{fig2}
\end{figure*}
\begin{enumerate}
   \item Firstly, a mobile two-hand reconstruction method based on lightweight attention mechanism named LWA-HAND (illustrated in Figure \ref{fig2}) is proposed and demonstrate the effectiveness of mobile attention for the two-hand reconstruction task.

   \item Secondly, a lightweight feature attention module to extract both local occlusion representation and global image patch representation is proposed in a coarse-to-fine manner, producing fusion of these two representation with low latency.

   \item Thirdly, a cross image and graph bridge module to fuse image context and hand vertex is proposed. It constructs a pyramid bridge to extract global context features of two hands with very few tokens in the attention mechanism, connecting context features directly to the hand vertex domain without extra transforms of calculation.

   \item Finally, we propose a lightweight cross attention mechanism which uses element-wise operation for cross attention of two hands with linear complexity \cite{mehta2022separable}, reducing the flops of attention operation in the representation of interacting hands.
\end{enumerate}

For the construction of interacting hands, our method reduces the calculation to $470M$ flops and achieves comparable results with existing solutions based on heavy-weight networks of $10$ times of flops on the InterHand 2.6M benchmark. This demonstrates the ability of the mobile transformers in the construction of interacting hands on resource-constrained devices.

\section{Related Work}
\subsection{Hand Reconstruction}
Hand reconstruction is studied for decades, including single hand reconstruction, two-hand reconstruction, and mesh regression. Some of the existing methods are already applied to virtual reality, robots and remote medical operations. The most related work of hand reconstruction is reviewed below.

\subsubsection{Single Hand Reconstruction}
Hand landmarks prediction and hand gesture recognition are well studied \cite{heap1996towards,wang2013video} before deep learning. The estimation of 3D hand landmarks from a single image is produced through deep learning networks \cite{cai2018weakly,mueller2018ganerated,spurr2018cross,zimmermann2017learning}. 
The {standard} parametric hand shape model MANO \cite{romero2022embodied} and a variety of large scale datasets \cite{joo2015panoptic,moon2020interhand2,simon2017hand,zimmermann2019freihand} are then available. Methods based on these are built to reconstruct the hand \cite{baek2019pushing,chen2021camera,ge20193d,kulon2020weakly,lin2021end,tang2021towards,zhang2021hand,zhou2020monocular} through deep networks. Among all  existing methods, the attention–based models \cite{lin2021end,lin2021mesh,li2022interacting} produce the best results as the representation of the global relationship of hand mesh vertices are learned well through the training of attention modules. However, this excellent performance relies on heavy weight attention mechanism which is impractical on mobile devices. Therefore, we propose our methods based on mobile transformers. 

\subsubsection{Two-Hand Reconstruction}
Although almost all of the above methods achieve good results on the single hand reconstruction task, interacting hands remains to be a challenge for the hand reconstruction task. First, previous methods simultaneously reconstruct body and hand \cite{choutas2020monocular,joo2018total,rong2020frankmocap,xiang2019monocular,zhang2021lightweight,zhou2021monocular}, each hand is treated separately and heavy occlusion of hand interaction could not be well represented. Second, a method based on a multi-view framework to reconstruct high-quality interactive hand \cite{smith2020constraining} could provide good results. However, its hardware setup is costly, the calculation is heavy energy consumption and it's relied on a variety number of cameras.
Third, other monocular tracking methods based on kinematic formulations are sensitive to fast motion, and tracking failures often occur using a depth sensor \cite{kyriazis2014scalable,mueller2019real,oikonomidis2012tracking,taylor2017articulated,tzionas2016capturing} or an RGB camera \cite{wang2020rgb2hands}. However, their dense mapping strategy relied on the correspondences between hand vertices and image representation is difficult to be processed. Then, deep learning based methods  \cite{fan2021learning,kim2021end,moon2020interhand2,rong2021monocular,zhang2021pymaf} are applied to reconstruct two-hand interaction through the application of a variety of feature maps. For example, they rely on 2.5D heatmaps to estimate hand joint positions \cite{fan2021learning,moon2020interhand2}, extract sparse image features \cite{xiao2018simple}, reconstruct each hand respectively and fine-tune for 3D landmark prediction later \cite{kim2021end,rong2020frankmocap}. However, the sparse representation of a single hand is not able to produce the efficient representation of hand surface occlusions and hands interaction context. Therefore, dense feature representation based on attention and graph \cite{li2022interacting} are studied to well learn the occlusion and context in the training. Unfortunately, the normal attention mechanism is computationally expensive and is hard to be deployed on mobile devices. In this paper, we propose several designs of lightweight attention modules to reduce the calculation and energy consumption. 

\subsection{Real-time Hand Reconstruction}
Currently, a variety of hand reconstruction methods are applied to mobile devices which require low latency and low energy consumption. In order to drive virtual and augmented reality (VR/AR) experiences on a mobile processor, a variety of hand reconstruction methods are proposed based on mobile network. For example, inverted residuals and linear bottlenecks \cite{sandler2018mobilenetv2} are used to build base blocks for hand pose, scale and depth prediction \cite{han2020megatrack}. Precise landmark localization of $21$ $2.5D$ coordinates inside the detected hand regions via regression is estimated to address CPU inference on the mobile devices \cite{zhang2020mediapipe}. However, these mobile methods based on mobile CNN blocks lack the ability of reconstruct hands with occlusions and interaction. The attention mechanisms and graph is not applied to build representation of local and global context of interacting hands. Therefore, we propose mobile attention and graph modules which can both handle challenging occlusions and interacting context with low flops.  

\section{Formulation}

\subsection{Two-Hand Mesh Representation}
Similar with the state-of-the-art models \cite{li2022interacting,moon2020interhand2,zhang2021interacting}, The popular parametric hand model MANO \cite{romero2022embodied} is adopted for each hand which contains $N = 778$ vertices. To utilize the mobile attention mechanism in the proposed modules, dense matching encoding for each vertex\cite{wang2020rgb2hands,li2022interacting} is used. As shown in Figure \ref{fig2}, our LWA-HAND has a hierarchical architecture that reconstructs hand mesh using a variety of coarse-to-fine blocks with different types such as mobile attention module, domain bridges between image context and hand vertex and pyramid hand vertex decoding modules \cite{li2022interacting}. To construct the coarse-to-fine mesh topology and enable the building of bridge between image context and hand representation in a coarse-to-fine manner, we apply the graph coarsening method in \cite{li2022interacting} and build $N$ $b = 3$ level submeshes with vertex number $N_{0} = 63$, $N_{1} = 126$, $N_{2} = 252$ and reserve the topological relationship between adjacent levels for upsampling. After the third block, a simple linear layer is employed to upsample the final submesh ($N_{2} = 252$) to the full MANO mesh ($N = 778$), producing the final two-hand vertices.

\subsection{Overview}

The proposed system contains three main parts: mobile vision attention encoder-decoder (green block and yellow blocks in Figure \ref{fig2}), pyramid attention-graph bridges (red, green and pink blocks in Figure \ref{fig2}),  and mobile interacting attention module (vertex representation in Figure \ref{fig2}). Given a single RGB image, a global feature vector $F_{G}$ is firstly produced through feeding it to an mobile vision attention encoder. Simultaneously, several bundled feature maps $\{ Y_{t} \in \mathbb{R}^{C_{t} \times H_{t} \times W_{t}}, t = 0,1,2\}$, where $t$ indicates that the $t$-th feature level corresponds to the input of the $t$-th bridge in the domain bridges, $H_{t} \times W_{t}$ is the resolution of the feature maps which gradually increases, and $C_{t}$ is the feature channel. The domain bridges take $Y_{t}, t=0,1,2$ as input and transform them to hidden features $Z_{t}, t=0,1,2$ through attention operation. Then, the bridges directly fuse global context features $Z_{t}, t=0,1,2$ with hand vertex features in a coarse-to-fine manner. Note that, at each lever, the stream runs through a vision mobile attention module, a domain bridge, and a graph decoded module. These modules are illustrated in Figure \ref{fig2} - Figure \ref{fig5} and will be discussed in Section $4.1$, Section $4.2$ and Section $4.3$, separately.   

\section{Light Former Graph}
Existing work of interacting hands reconstruction \cite{li2022interacting,zhang2021interacting,moon2020interhand2} is built with convolutional encoder and attention module with lots of energy consumption. In this paper, we propose a lightweight former graph architecture  as shown in Figure \ref{fig2} to represent the interaction and occlusion of two-hands. The Light former graph is consisted of three main modules based on lightweight blocks including lightweight feature attention module, pyramid cross image and graph bridge module and lightweight cross hand attention module. They are introduced in the following.

\subsection{Lightweight Feature Attention Module}
\begin{figure*}
\centering
\includegraphics[height=2cm]{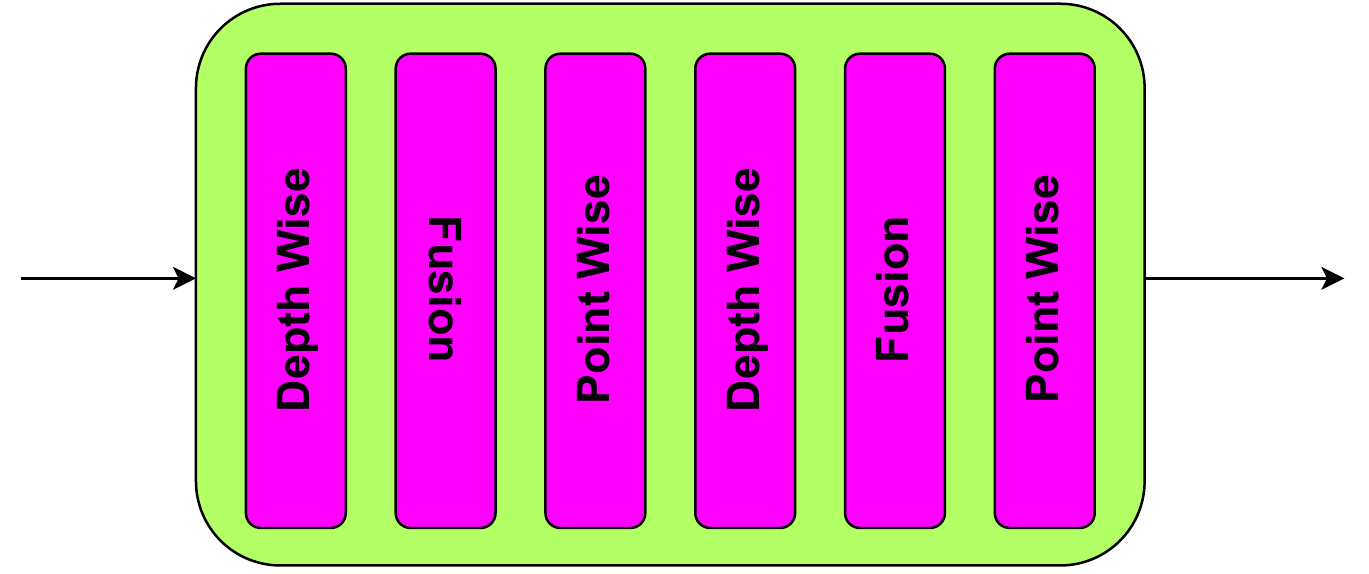}
\caption{Mobile module with attention.}
\label{fig3}
\end{figure*}
Lightweight convolutional neural networks (CNNs) such as MobileNets \cite{howard2019searching,howard2017mobilenets,sandler2018mobilenetv2} efficiently encode local features by stacking depthwise and pointwise convolutions. Mobile vision transformers and its follow-ups \cite{dong2022cswin,liu2021swin,touvron2021training} achieve global features through tokens. Inspired by the above advantages, a recent mobile architecture \cite{chen2022mobile} is applied to connect local features and global features as shown in Figure \ref{fig3}.

This module takes an image as the first input $\{ \textbf{X} \in \mathbb{R}^{(H \times W \times 3)} \}$ and applies inverted bottleneck blocks \cite{sandler2018mobilenetv2} to extract local features. Besides, learnable parameters (tokens) $\textbf{Z} \in \mathbb{R}^{M \times d}$ are taken as the second input where $M$ and $d$ are the number and dimension of tokens, respectively. Similar to \cite{li2022interacting},  these tokens are randomly initialized, and a small number of tokens ($M < 7$) is applied to represent a global prior of the image. Therefore, the operation of  inverted bottleneck blocks and tokens results in much less computational effort.

In order to make fusion of the global and local features in the encoder, the lightweight encoding is computed by:

\begin{equation}
    F_{X_{i}^{0}} = H(F_{X_{i-1}^{1}},Z_{i}) = \text{Fusion}[\text{CNN}_{\text{Depth-wise}}(F_{X_{i-1}^{1}}),Z_{i}],
\end{equation}

\begin{equation}
    F_{X_{i}^{1}} = H(F_{X_{i}^{0}}) = \text{CNN}_{\text{Point-wise}}(F_{X_{i}^{0}}),
\end{equation}
where $i=0,1,2,3,...,N_{\text{stack}}$, $N_{\text{stack}}$ is the number of stacks, $X_{i}^{0},X_{i}^{1}, i > 0$ are local features of the two stages in the $i$-th stack, and $X_{0}=X_{0}^{0}=X_{0}^{1}$ represents the input image. Here $Z_{i}$ represents the global features in the $i$-th stack, $\text{CNN}_{\text{Depth-wise}}$ represents the depth wise convolution, $\text{CNN}_{\text{Point-wise}}$ represents the point wise convolution, $\text{Fusion}$ represents the contact operation. $F_{X_{i}^{0}}$ represents the feature of the stage $0$ at the $i$-th stack, $F_{X_{i}^{1}}$ represents the local feature of the stage $1$ at the $i$-th stack, $H(X_{i-1}^{1},Z_{i})$ represents the convolutional functions with the input $F_{X_{i-1}^{1}}$ and $Z_{i}$. The fusion operation directly contact global tokens $Z_{i}$ with local features $\text{CNN}_{\text{Depth-wise}}(X_{i-1}^{1})$ rather than other attention mechanism with heavy calculation. 

\subsection{Pyramid Cross Image and Graph Bridge Module}
\begin{figure*}
\centering
\includegraphics[height=4cm]{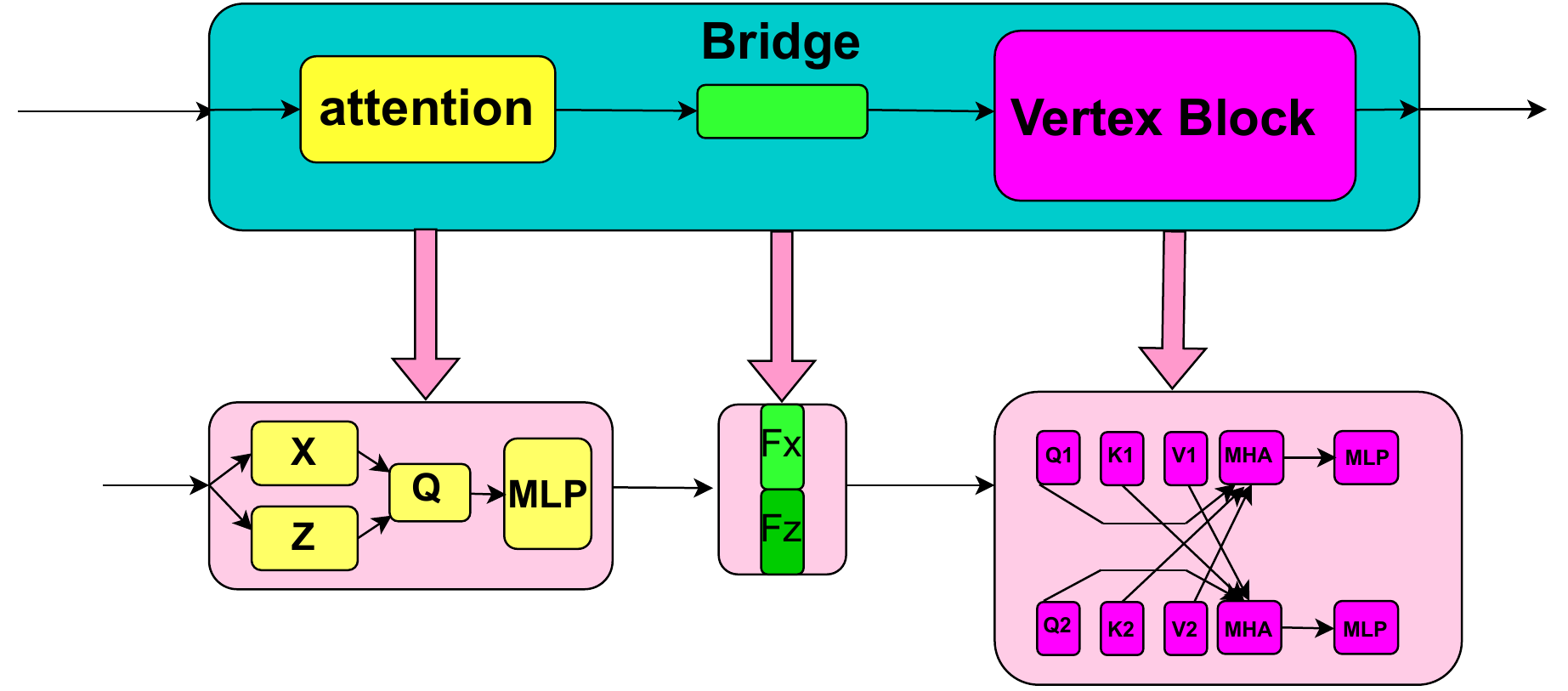}
\caption{Graph bridge module. The yellow block is the attention module for building an attention map between the local feature $X$ and the global feature $Z$ through an attention matrix $Q$ and $\text{MLP}$ module. The fusion module (in green) is the direct contact module for the feature $Fx$ and $Fz$. The pink vertex module is built with cross hand attention module with two $Q$ matrix $Q_{1}$ and $Q_{2}$, two key matrix $K_{1}$ and $K_{2}$, two value matrix $V_{1}$ and $V_{2}$.}
\label{fig4}
\end{figure*}
Communication between CNN and transformer through a bridge is an efficient way to produce fusion of different domain representation \cite{mehta2021mobilevit,chen2022mobile}. Moreover, communication between image and graph is demonstrated as an efficient way to feed context features into the representation of hand vertex. The pyramid cross image and graph bridge module build two types of communication: communication between local features and global tokens and communication between global features and hand vertex. Furthermore, the communication between the triple domains is built in a coarse-to-fine manner (see, Figure \ref{fig4}): Firstly, unlike the two-way bridge to connect local and global features \cite{chen2022mobile}, the local features and global features are communicated through an one-way bridge. A lightweight cross attention is  applied where $(\textbf{W}^{Q}, \textbf{W}^{K}, \textbf{W}^{V})$ are the matrices of three projections, only $\textbf{W}^{Q}$ is used. Specifically, the lightweight cross attention from local features map $\textbf{X}$ to global tokens $\textbf{Z}$ is computed by:
\begin{equation}
    A_{F_{X_{i}^{1}} \rightarrow Z_{i}}  =  [\text{Attn}(\bar{z}_{i}\textbf{W}_{i}^{Q},\bar{x}_{i},\bar{x}_{i})]_{i=1:h} \textbf{W}^{O},
\end{equation}
where the local feature $F_{X_{i}^{1}}$ and global tokens $Z_{i}$ are split into $h$ heads as $F_{X_{i}^{1}} = [\bar{x}_{1},...,\bar{x}_{h}] $, $Z_{i} = [ \bar{z}_{1},...,\bar{z}_{h} ]$ for multi-head attention. The split for the $i$-th head is represented as $\bar{z}_{i} \in \mathbb{R}^{M \times \frac{d}{h}}$, $W_{i}^{Q}$ is the query projection matrix for the $i$-th head, $and W_{o}$ is used to combine multiple heads together. 

Secondly, at the $i$-th stack, the global features $Z_{i}$ are  mapped to the domain of graph representation in the bridge, denoted as $M_{i}$. Here, $M_{i}$ is computed by:
\begin{equation}
    M_{i} = A_{F_{X_{i}^{1}} \rightarrow Z_{i}} F_{X_{i}^{1}}
\end{equation}
where $A_{F_{X_{i}^{1}} \rightarrow Z_{i}}$ is the attention matrix calculated previously. Here $F_{X_{i}^{1}}$ is the local feature at the $i$-th stack and $Z_{i}$ is the global feature at the $i$-th stack.

The mapping is calculated in two stages: direct connection stage and  cross attention stage. At the first stage, the global feature $M_{i}$ is directly contacted with hand vertex representation $V_{i}$ at the $i$-th stage for each hand. At the second stage, a cross attention mechanism is applied between two hands \cite{li2022interacting} for deep fusion. The two-stage fusion is calculated as the following:
\begin{equation}
    V_{i}^{Ro} = \text{Fusion}(V_{i}^{R},M_{i}),\quad V_{i}^{Lo} = \text{Fusion}(V_{i}^{L},M_{i}),
\end{equation}
    \begin{equation}
    FH_{i}^{R \rightarrow L} = \text{softmax} \left(\frac{f(Q_{i}^{Lo})f(K_{i}^{Ro})}{d}\right)f(V_{i}^{Ro}),
\end{equation}
\begin{equation}
    FH_{i}^{L \rightarrow R} = \text{softmax} \left(\frac{f(Q_{i}^{Ro})f(K_{i}^{Lo})}{d}\right)f(V_{i}^{Lo}),
\end{equation}
where at the $i$-th stack, $V_{i}^{R}$ and $V_{i}^{L}$ are the representations of right hand vertex and left hand vertex before fusion, $V_{i}^{Ro}$ and $V_{i}^{Lo}$ are the representations of  right hand vertex and  left hand vertex after fusion, the operation of direct contact is applied to make the fusion. Multi-head self-attention (MHSA) module is applied to obtain the query, key and value features of each hand representation $V_{i}^{ho}, h \in {L, R}$ after the fusion, and the values are indicated by $Q_{i}^{ho}$, $K_{i}^{ho}$, $V_{i}^{ho} , h \in {L,R}$.  $T$ represents the matrix transpose. Like interacting hands in the existing methods \cite{li2022interacting}, $FH_{i}^{R \rightarrow L}$ and $FH_{i}^{L \rightarrow R}$ are the cross hand attention features encoding the correlation between two hands.  $d$ is a normalization constant and $f$ represents the function with the three features as input respectively. The cross-hand attention features are merged into the hand vertex features by a pointwise MLP layer $fp$ as
\begin{equation}
    FH_{i}^{La} = fp(FH_{i}^{Lo} + FH_{i}^{R \rightarrow L}),
\end{equation}
\begin{equation}
    FH_{i}^{Ra} = fp(FH_{i}^{Ro} + FH_{i}^{L \rightarrow R}),
\end{equation}
where $FH_{i}^{La}$ and $FH_{i}^{Ra}$ are the output hand vertex features at the $i$-th stack, which act as the input of both hands at the next stack.

\subsection{Lightweight Cross Hand Attention Module}
\begin{figure*}
\centering
\includegraphics[height=5.5cm]{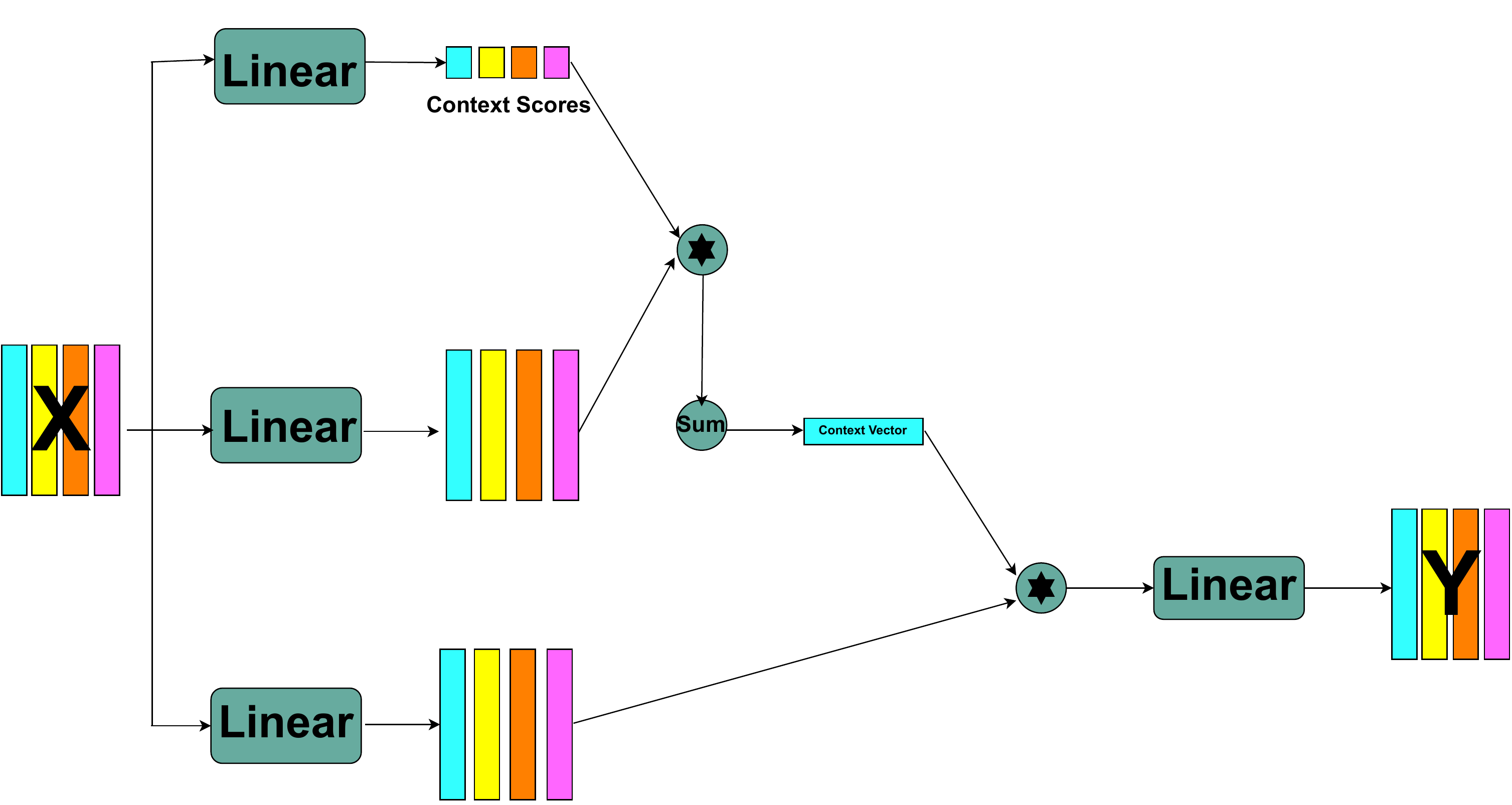}
\caption{Cross hand attention module.}
\label{fig5}
\end{figure*}
In order to reduce the calculation of the above mentioned cross attention between two hands, a separable self-attention structure \cite{mehta2022separable} is applied to construct a lightweight cross hand attention module. The input $FH_{i}^{h}, h \in {L,R}$ is processed using three branches: input $I$, key $K$ and value $V$ as shown in Figure \ref{fig5}.

Firstly, the input branch $I$ is responsible to calculate the value of query features $Q_{i}^{ho}, h \in {L,R}$. The input branch $I$ maps each $d$ dimensional token in $FH_{i}^{ho}, h \in L,R$ to a scalar using a linear layer with weights $W_{i}^{I} \in \mathbb{R}^{d}$. This linear projection is an inner-product operation and computes the distance between latent token $L$ and $FH_{i}^{h}$, resulting in a $k$-dimensional vector. A softmax operation is then applied to this $k$ dimensional vector to produce context scores $c_{s} \in \mathbb{R}^{k}$ representing the value of query features.

Secondly, the context scores $c_{s}$ are used to compute a context vector $c_{v}$. Specifically, the input $FH_{i}^{h}$ is linearly projected to a $d$ dimensional space using $K$ branch with weights $W_{k} \in \mathbb{R}^{d \times d}$ to produce an output $FH_{ik}^{h} \in \mathbb{R}^{k \times d}$ representing the value of key features. The context vector $c_{v} \in \mathbb{R}^{d}$ is then computed as a weighted sum of $FH_{ik}^{h}$:
\begin{equation}
    c_{v} = \sum_{i=1}^{k} c_{s}(i)FH_{ik}^{h}(i).
\end{equation}

Finally the contextual information encoded in $c_{v}$ is shared with all tokens in $FH_{i}^{h}$. The input $FH_{i}^{h}$ is linearly projected to a $d$ dimensional space using the value branch $V$ with wights $W_{v} \in \mathbb{R}^{d \times d}$ to produce an output $FH_{iv}^{h}$ representing the value of the value features. 

\subsection{Loss Function}
To train the proposed mobile network and ensure a fair comparison with the state-of-the-art models \cite{li2022interacting,moon2020interhand2,zhang2021interacting}, vertex loss, regressed joint loss and mesh smooth loss are utilized. They appear in IntagHand \cite{li2022interacting}, and other loss items such as segmentation loss are included in order to demonstrate the efficiency of our proposed mobile network architecture.

\section{Experiment}

\subsection{Experimental Settings}
\subsubsection{Implementation Details}
Our network is implemented using PyTorch. The proposed network with efficient modules are trained in an end-to-end manner.

\subsubsection{Training Details}
For a fair comparison, data augmentation techniques such as scaling, rotation, random horizontal flip and color jittering are used in the training of the proposed network and the state-of-the-art models \cite{li2022interacting,zhang2021interacting,moon2020interhand2}.
We use Adam optimizer \cite{kingma2014adam}. The training is carried on $6$ NVIDIA RTX $2080$Ti GPUs. The minibatch size is set to $32$. We train $120$ epochs and it takes us one day for training the network. The learning rate decays to $1\times10^{-5}$ at the $50$-th epoch from the initial rate $1\times10^{-4}$.

\subsubsection{Evaluation Metrics}
In fair comparison with other state-of-the-art models \cite{li2022interacting,zhang2021interacting,moon2020interhand2}, the mean per joint position error (MPJPE) and mean per vertex position error (MPVPE) in millimeters are calculated. Besides, we follow standards of Two-Hands \cite{zhang2019end}, the length of the middle metacarpal of each hand is scaled to $9.5cm$ for training, and the length of the middle metacarpal of each and is re-scaled back to the ground truth bone length during evaluation. This operation is performed after root joint alignment of each hand.

\subsection{Datasets}
\subsubsection{InterHand2.6M Dataset \cite{moon2020interhand2}}
All networks in this paper are trained on InterHand2.6M \cite{moon2020interhand2}, this dataset is the only dataset with two-hand mesh annotation. In detail, the interacting two-hand (IH) data with both human and machine (H+M) annotated is applied for the two-hand reconstruction evaluation. The training samples are $366K$ and the testing samples are $261K$. At preprocessing, the hand region is cropped out according to the $2D$ projection of hand vertices. The resolution of each input image is set as $256 \times 256$ resolution \cite{li2022interacting}.

\subsection{Quantitative Results}
\begin{figure*}
\centering
\includegraphics[height=4.5cm,width=12cm]{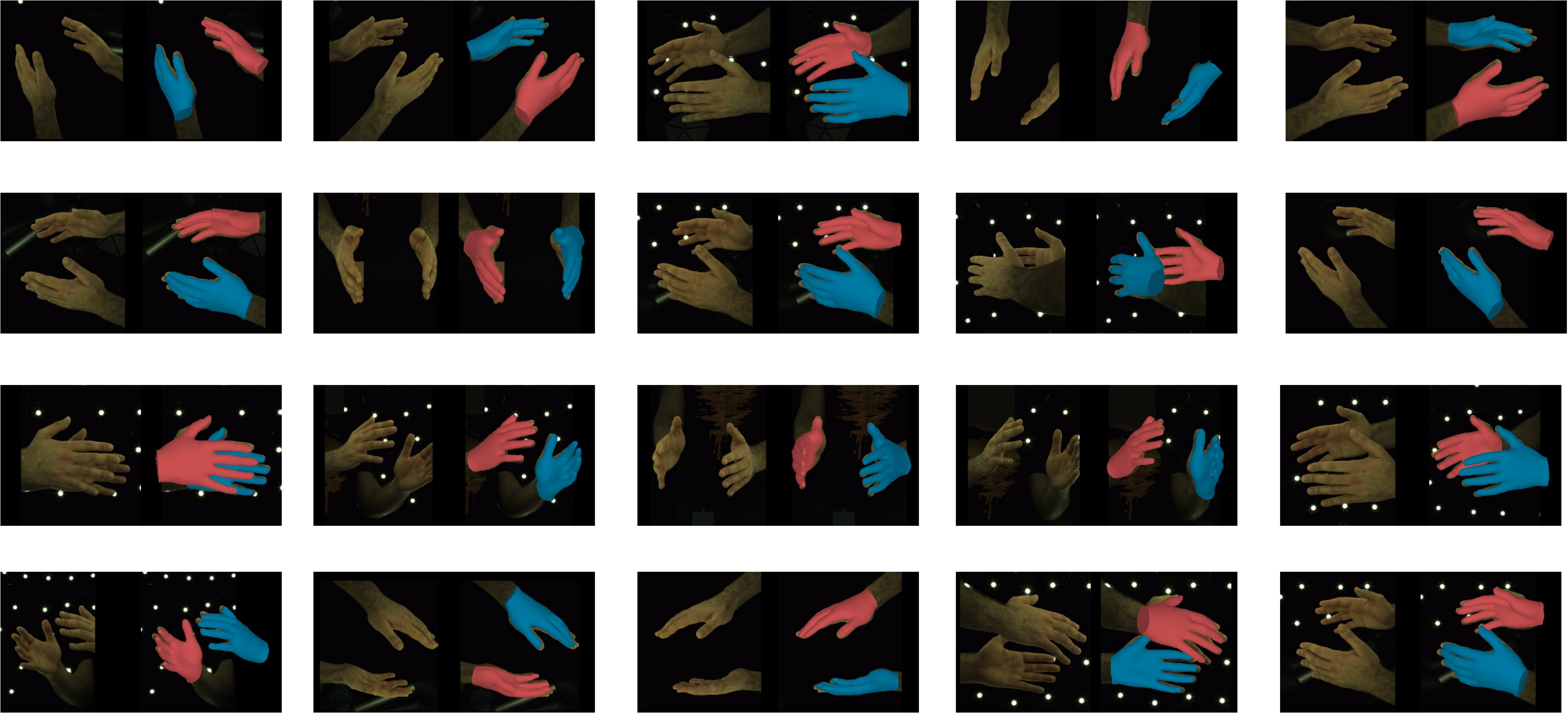}
\includegraphics[height=4.5cm,width=12cm]{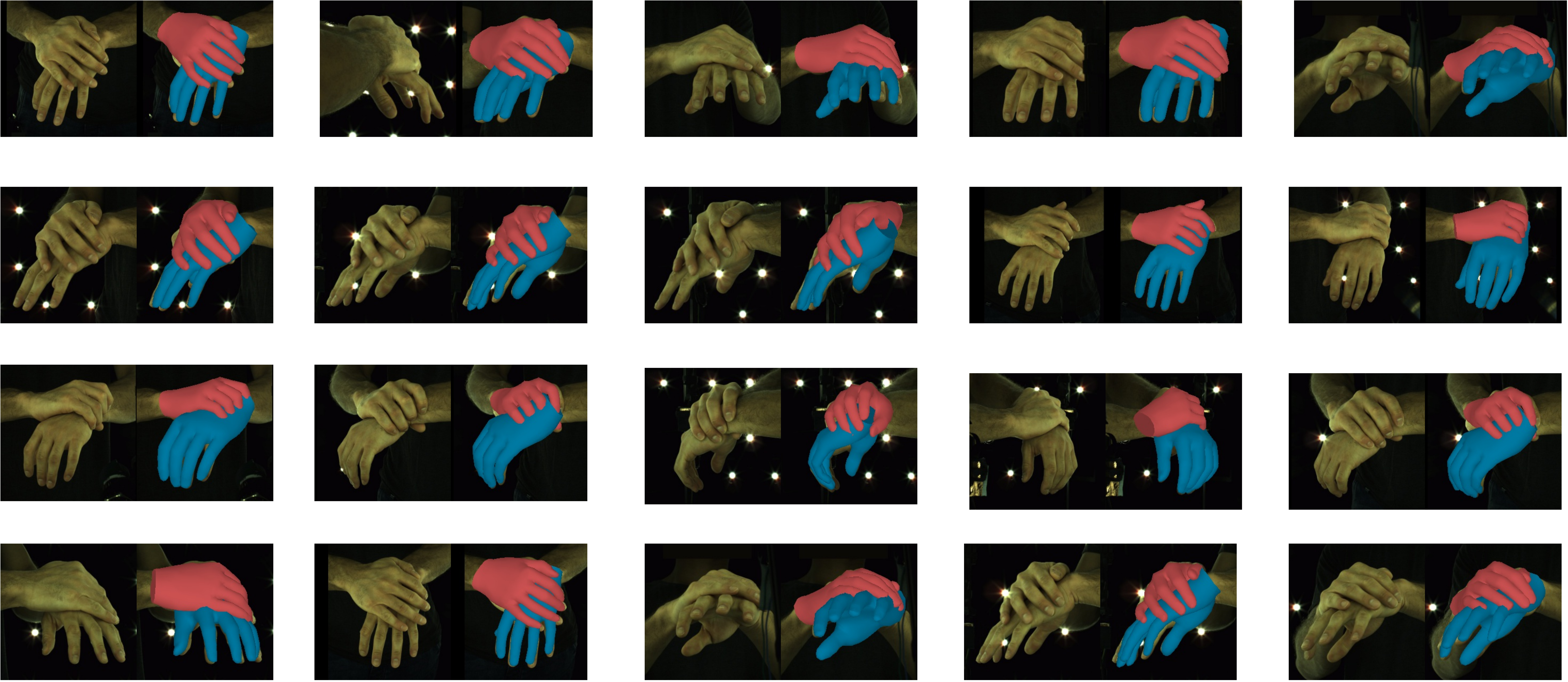}
\includegraphics[height=4.5cm,width=12cm]{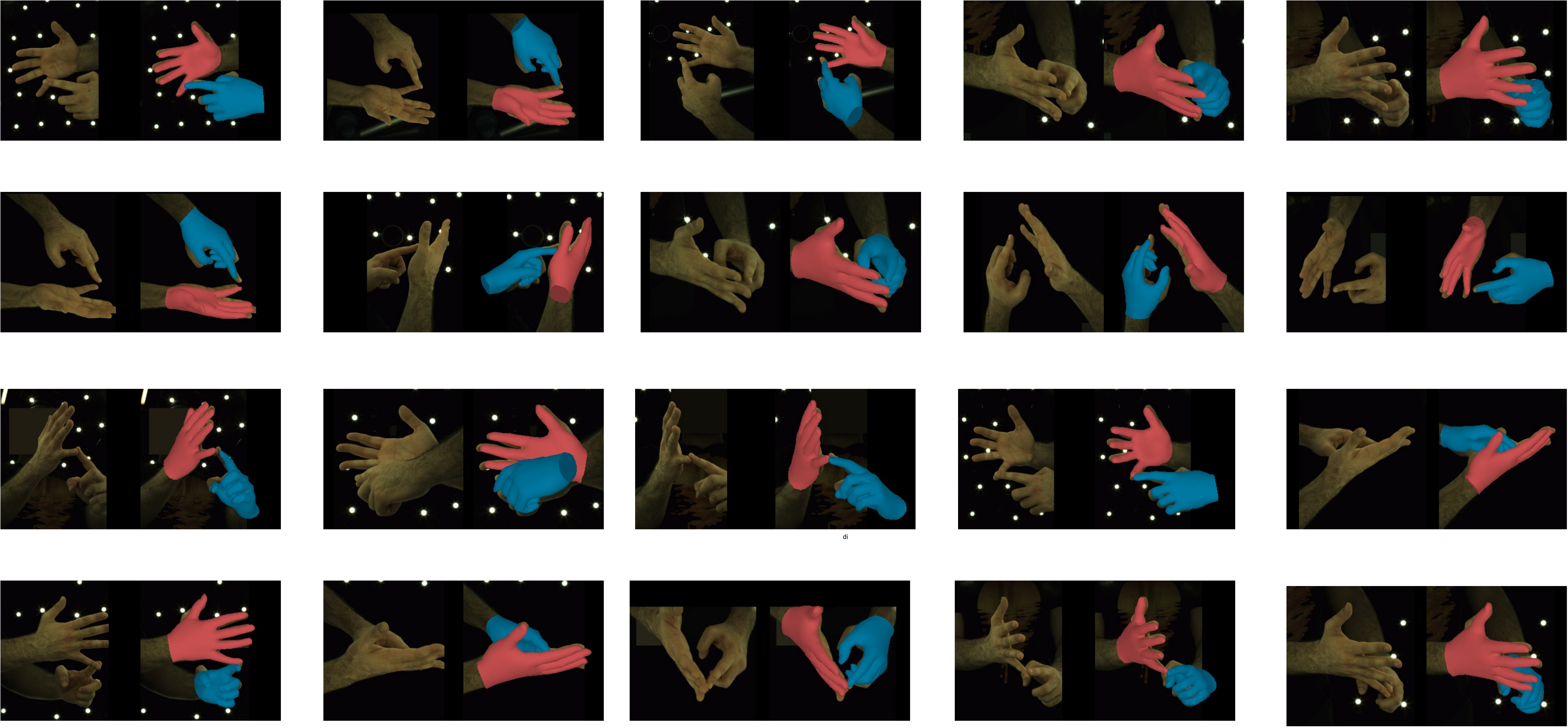}
\includegraphics[height=4.5cm,width=12.2cm]{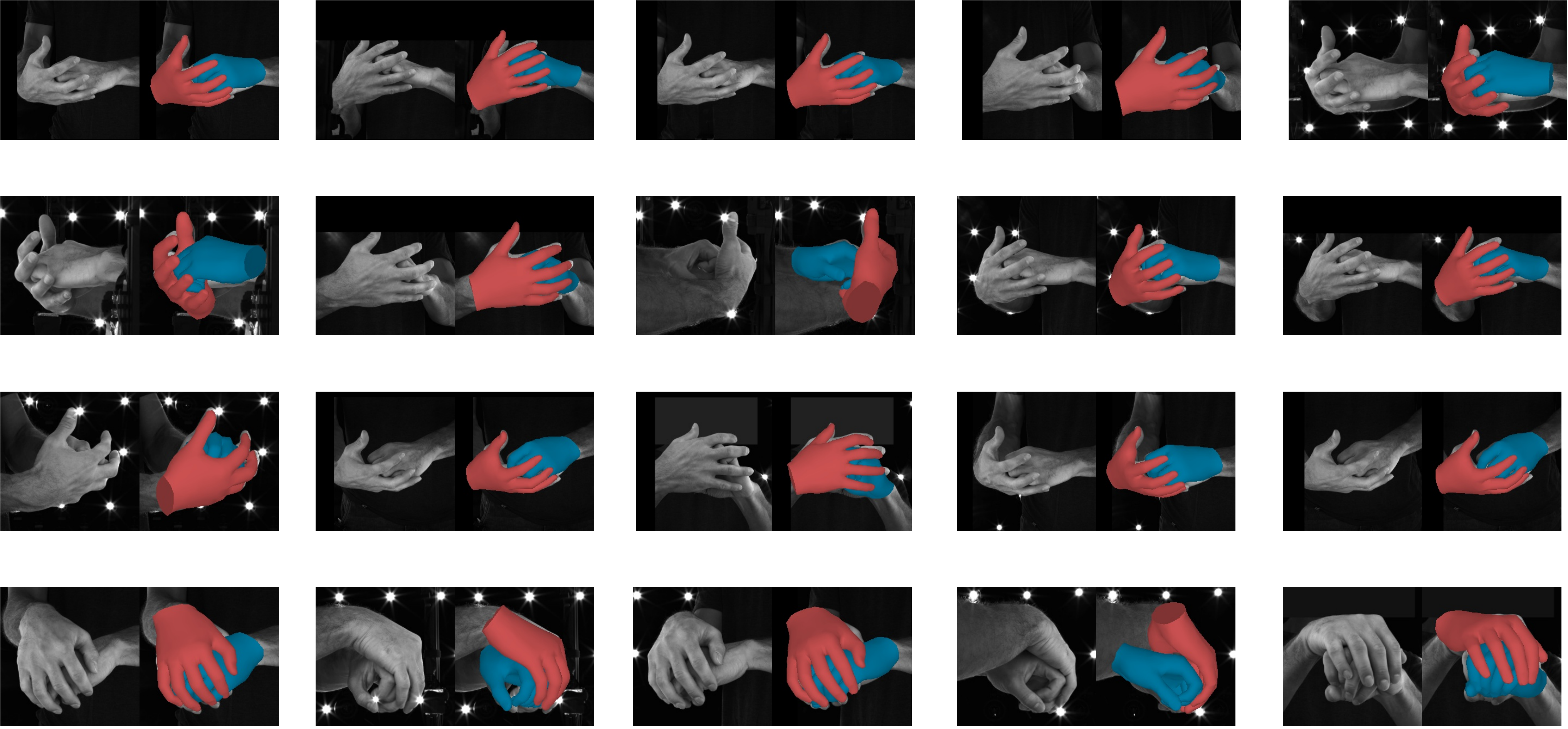}
\caption{Illustrations for interacting hand reconstruction by our proposed model.}
\label{fig9}
\end{figure*}

We first compare our proposed mobile network with the state-of-the-art two-hand reconstruction methods and recent two-hand reconstruction methods. The first model \cite{moon2020interhand2} regresses 3D skeletons of two hands directly. The second model \cite{zhang2021interacting} predicts the pose and shape parameters of two MANO \cite{romero2022embodied} models. The third model \cite{li2022interacting} predicts the vertex of interacting-hands with graph and attention modules. For a fair comparison, we run their released source code on the same subset of Inter-Hand2.6M \cite{moon2020interhand2}. Comparison results are shown in Table \ref{table1}. Figure \ref{fig9} shows the results of our proposed model. It is clearly shown in Table \ref{table1} that our method achieved comparable MPJPE and MPVPE as the state-of-the-art models. Furthermore, we reduces the flops to $0.47GFlops$ while the flops of the state-of-the-art models are around $10GFlops$ or higher value. 

\begin{table*}[h]
\caption{Comparison with the state-of-the-art models on performance and flops. The mean per joint position error (MPJPE) and mean per vertex position error (MPVPE) are calculated in millimeters. The flops are calculated in GFlops.}
\begin{center}
\begin{tabular}{|c|c|c|c|c|c|}
\hline

\hfil Model    &\hfil MPJPE  &\hfil MPVPE  & \hfil Total flops & \hfil Image part  & \hfil Pose part  \\\hline

\hfil Inter-Hand \cite{moon2020interhand2}    &$16.00$ &$-$ &$19.49$  &$5.37$ &$14.12$\\\hline

\hfil Two-Hand-Shape \cite{zhang2021interacting}   &$13.48$ &$13.95$ &$28.98$  &$9.52$ &$19.46$\\\hline

\hfil Intag-Hand \cite{li2022interacting}  &$8.79$ &$9.03$  &$8.42$ &$7.36$ &$1.06$\\\hline

\hfil Ours    &$12.56$ &$12.37$ &$0.47$  &$0.25$ &$0.22$\\\hline

\end{tabular}
\label{table1}
\end{center}
\end{table*}

\section{Discussion}
\subsection{Conclusion}
We present the proposed mobile method to reconstruct two interacting hands from a single RGB image. In this paper, we first introduce a lightweight feature attention module to extract both local occlusion representation and global image patch representation in a coarse-to-fine manner.  We next propose a cross image and graph bridge module to fuse image context and hand vertex. Finally, we  propose a lightweight cross attention mechanism which uses element-wise operation for cross attention of two hands. Comprehensive experiments demonstrate the comparable performance of our network on InterHand2.6M dataset, and verify the effectiveness and practicability of the proposed model in the real-time applications with low flops.

\subsection{Limitation}
The major limitation of our method is the high MPJPE and MPVPE. The proposed mobile modules reduce the flops while increase the error at the same time. We are studying efficient methods to further reduce the error for mobile application of two-hand reconstruction. 

\bibliographystyle{splncs04}
\bibliography{egbib}
\end{document}